\documentclass{article}
\usepackage{spconf,amsmath,amssymb,graphicx,hyperref}
\usepackage{enumitem}
\usepackage{multirow}
\usepackage{multicol}
\usepackage{threeparttable}
\usepackage{booktabs}
\usepackage{xcolor}
\usepackage{pifont}

\newcommand{\best}[1]{\textbf{#1}}
\newcommand{\second}[1]{\underline{#1}}

\title{Distilling Time Series Foundation Models for Efficient Forecasting}
\name{Yuqi Li \textsuperscript{a} $^{\dagger}$, Kuiye Ding \textsuperscript{a} $^{\dagger}$, Chuanguang Yang \textsuperscript{b}, Szu-Yu Chen\textsuperscript{c}, Yingli Tian\textsuperscript{a} $^{\star}$ \thanks{$^{\dagger}$ Equal contribution; $^{\star}$ Corresponding author: ytian@ccny.cuny.edu.}}
\address{\textsuperscript{a} The City College of New York, City University of New York, USA\\
  \textsuperscript{b} Institute of Computing Technology, Chinese Academy of Sciences, China\\
  \textsuperscript{c} Stevens Institute of Technology, USA\\
  }
\begin{document}
\ninept
\maketitle

\begin{abstract}
Time Series foundation models (TSFMs) deliver strong forecasting performance through large-scale pretraining, but their large parameter sizes make deployment costly. While knowledge distillation offers a natural and effective approach for model compression, techniques developed for general machine learning tasks are not directly applicable to time series forecasting due to the unique characteristics. To address this, we present \textbf{DistilTS}, the first distillation framework specifically designed for TSFMs. DistilTS addresses two key challenges: (1) \textit{task difficulty discrepancy}, specific to forecasting, where uniform weighting makes optimization dominated by easier short-term horizons, while long-term horizons receive weaker supervision; and (2) \textit{architecture discrepancy}, a general challenge in distillation, for which we design an alignment mechanism in the time series forecasting. To overcome these issues, DistilTS introduces horizon-weighted objectives to balance learning across horizons, and a temporal alignment strategy that reduces architectural mismatch, enabling compact models. Experiments on multiple benchmarks demonstrate that DistilTS achieves forecasting performance comparable to full-sized TSFMs, while reducing parameters by up to \textbf{1/150} and accelerating inference by up to \textbf{6000×}. Code is available at: \url{https://github.com/itsnotacie/DistilTS-ICASSP2026}.


\end{abstract}

\section{Introduction}
\label{sec:intro}

Time series forecasting is a central problem in the time series community~\cite{icassp1, qiu2025duet, HOU2026108140,li2025energypatchtst,li2025timeflowdiffuser}, where the goal is to predict future values from past observations and their temporal dynamics~\cite{icassp2, icassp3, qiu2024tfb,li2025frequency}. The field has recently entered the era of \emph{time series foundation models} (TSFMs), with parameter counts already exceeding billions~\cite{shi2024timemoe} and new architectures emerging in rapid succession~\cite{woo2024moirai,ansari2024chronos}. These models demonstrate strong forecasting performance and show clear advantages in zero-shot scenarios. However, their massive parameter sizes and limited inference efficiency present serious barriers to deployment in practice. This situation calls for a TSFM distillation framework that preserves the generalization ability of TSFMs while enabling efficient deployment.

Current approaches to knowledge distillation for time-series forecasting follows two primary directions: 1) \textbf{LLM-based distillation.} Recent studies leverage LLMs within time-series forecasting paradigms, either as direct predictors~\cite{jin2023time, gruver2023llmtime} or as feature enhancers~\cite{liu2024timecma}. In this paradigm, the teacher model remains an LLM, and the distillation process largely follows practices from LLM knowledge distillation, such as leveraging privileged information or correlation-based guidance, to train a smaller student model~\cite{liu2025efficientmultivariatetimeseries,sun2025timemkgknowledgeinfusedcausalreasoning}. 2) \textbf{Lightweight Model-based distillation.} Another line distills across conventional forecasting models, such as Transformers or CNNs into lightweight MLP students~\cite{ni2025timedistillefficientlongtermtime,yang2022cross,yang2024clip}. These approaches improve efficiency by transferring temporal or frequency-domain patterns but the distilled targets are either large language models or small-parameter forecasting models, rather than time series foundation models. To date, no work has specifically addressed knowledge distillation for time-series foundation models (TSFMs), whose billion-scale parameters pose unique challenges for compression, scalability, and real-world deployment.

Distilling TSFMs, however, is far from straightforward, as forecasting poses unique challenges not typically encountered in other fields. Two issues are particularly critical: 1) \textbf{Task difficulty discrepancy}: short-term forecasting is easier and yield smaller errors, while long-term forecasting is significantly more difficult due to increasing uncertainty over time. According to the \emph{seesaw effect mechanism} in multi-task learning~\cite{fu-etal-2025-training}, optimization naturally favors the easier objectives and neglects the harder ones. This imbalance is also evident in empirical results, where forecasting errors (e.g., MAE) grow substantially with horizon length, as shown in Table~3 of~\cite{AI}. 2) \textbf{Architecture discrepancy}: Structural mismatches between teacher and student models can significantly affect distillation performance~\cite{ijcv/LiuCLHDLM24}. In our design, we use both MLP and Transformer-based student models. For the Transformer, we adopt iTransformer~\cite{liu2023itransformer} because its variate-wise embeddings are much lighter than patch-level~\cite{woo2024moirai} or point-wise~\cite{chen2025a, shi2024timemoe} encodings. However, most TSFMs employ patch-level or point-wise states, making it difficult to directly align hidden representations between teacher and student during distillation.


\begin{figure*}[t]
  \centering
  \includegraphics[width=1\linewidth]{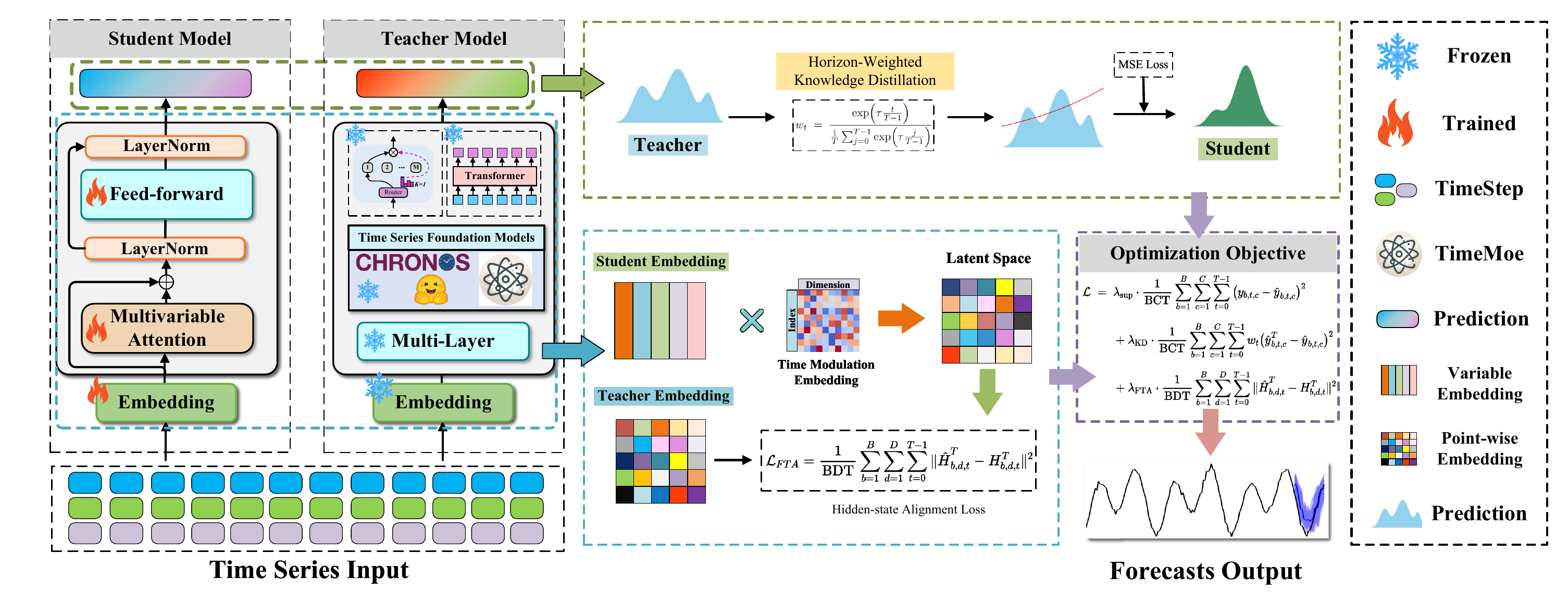}
  \vspace{-25pt}
  \caption{Overview of DistilTS. Horizon-weighted objectives rebalance supervision across horizons, and a factorized temporal alignment module projects variate-wise student embeddings into the teacher’s point-wise space.}
  \vspace{-10pt}
  \label{fig:framework}
\end{figure*}

To overcome these challenges, we present \textbf{DistilTS}, the first distillation framework specifically designed for TSFMs. DistilTS is built on two key components. First, \textbf{horizon-weighted objectives} redistribute the training signal, ensuring that long-horizon forecasts receive sufficient attention and are not overshadowed by easier short-term tasks. Second, a \textbf{factorized temporal alignment} module bridges the architectural gap between teacher and student models by projecting the student’s variate-wise embeddings into the teacher’s point-wise representation space, thus facilitating more effective alignment of hidden states. Together these designs yield lightweight distilled models that maintain the generalization ability of TSFMs and achieve stronger long-horizon fidelity. Compare to TSFMs which provide high forecasting capacity at the cost of heavy computation, our DistilTS delivers comparable performance with significantly more efficient inference. In this work, we make the following contributions:
\begin{itemize}
    \item We introduce \textbf{DistilTS}, the first distillation framework specifically designed for TSFMs, explicitly addressing the unique challenges of task difficulty discrepancy and architecture discrepancy.
    \item We propose two key innovations: 
(\emph{i}) horizon-weighted objectives that address task difficulty discrepancy by redistributing training signals across across different forecasting horizons to ensure balanced learning, and 
(\emph{ii}) a factorized temporal alignment module that addresses architecture discrepancy by projecting student embeddings into the teacher’s temporal space to enable more effective knowledge transfer.
    \item Experiments on real-world benchmarks demonstrate that DistilTS achieves performance comparable to full-scale TSFMs while significantly improving efficiency. We also release an open-source framework covering representative TSFMs.
\end{itemize}

\section{Methodology}
\label{sec:method}

\subsection{DistilTS Framework}
\label{sec:framework}

DistilTS introduces two complementary innovations that together enable effective distillation for TSFMs: \emph{horizon-weighted objectives} and \emph{factorized temporal alignment}, as summarized in Figure~\ref{fig:framework}. These components are designed to resolve two key challenges in time series distillation: the \textbf{task difficulty discrepancy}, where optimization favors short horizons over long ones, and the \textbf{architecture discrepancy}, where student and teacher adopt different representational structures.


\subsubsection{Horizon-weighted knowledge distillation}

Classic knowledge distillation methods are mainly developed for classification tasks, where supervision does not explicitly distinguish temporal
structure. Directly averaging errors across prediction steps yields an unbiased
empirical risk estimate, but in practice optimization is dominated by the easier
short horizons. Short-term forecasting exhibit stronger and more consistent
gradients, while long-term forecasting suffer from lower signal-to-noise ratios
and higher variance. This imbalance leads to under-training of long horizons,
even for models with sufficient capacity.

To address this optimization imbalance, we introduce an exponentially
increasing weighting scheme across horizons. Later steps are assigned larger
weights, which enhances the effective gradient signal for long-term forecasting
and improves fidelity under limited computation. The weights are defined as
\begin{equation}
w_t = \frac{\exp\!\Big(\tau \tfrac{t}{T-1}\Big)}
{\tfrac{1}{T}\sum_{j=0}^{T-1}\exp\!\Big(\tau \tfrac{j}{T-1}\Big)}, \quad t=0,\dots,T-1 .
\end{equation}
The supervised and distillation losses then become
\begin{equation}
\mathcal{L}_{\text{sup}} = 
\frac{1}{\mathrm{BCT}}
\sum_{b=1}^B \sum_{c=1}^C \sum_{t=0}^{T-1}
\big(y_{b,t,c} - \hat{y}_{b,t,c}\big)^2 ,
\end{equation}

\begin{equation}
\mathcal{L}_{\text{KD}} = 
\frac{1}{\mathrm{BCT}}
\sum_{b=1}^B \sum_{c=1}^C \sum_{t=0}^{T-1}
w_t \big(\hat{y}^{T}_{b,t,c} - \hat{y}_{b,t,c}\big)^2.
\end{equation}

This design shares a conceptual analogy with the temperature parameter in classification-based knowledge distillation. In that setting, a softened softmax distribution exposes informative signals from less confident classes. In our case, the horizon parameter $\tau$ amplifies the contribution of harder long-horizon steps, thereby balancing the optimization signal across time. In the idealized case of unlimited data and computation, uniform averaging would suffice. However, in practical settings horizon-aware weighting allows the long-range strengths of foundation models to be distilled more effectively into lightweight student models.

\subsubsection{Factorized Temporal Alignment}

A key challenge in distillation arises from the architectural mismatch between teacher and student representations. The teacher adopts a point-wise design, producing point-wise hidden states $H^{T}\!\in\!\mathbb{R}^{B\times D\times T\times d_T}$,
where each variate is represented by a sequence of embeddings across horizon steps. In contrast, the student follows the variate-wise design of iTransformer, encoding each variate into a single embedding $H^{S}\!\in\!\mathbb{R}^{B\times D\times d_S}$. This choice is motivated by efficiency: patch-based tokenization segments each variate into multiple local embeddings, inflating the token count and parameter size, whereas variate-wise tokenization compresses temporal information into one embedding per variate, yielding a much lighter student model. The resulting discrepancy is that teacher states are indexed by both variate and time, while student states only capture variate-wise information.

To enable effective hidden transfer under this mismatch, we propose a
\emph{factorized temporal alignment} module. Each student variate embedding is
first projected into a latent space, then modulated by a learnable time
embedding, and finally mapped into the teacher space:
\begin{equation}
\hat{H}^{T}_{b,d,t} = W_{\mathrm{out}}\, 
\phi\!\left( (W_s h^{S}_{b,d}) \odot E_t \right),
\end{equation}
where $h^{S}_{b,d}$ denotes the student’s embedding of variate $c$ in sample $b$. Here, $c$ indexes the original input channel, while $d$ indexes the corresponding variate-wise embedding after encoding, 
with a one-to-one correspondence between $c$ and $d$.

$W_s\!\in\!\mathbb{R}^{d_S\times u}$ and 
$W_{\mathrm{out}}\!\in\!\mathbb{R}^{u\times d_T}$ are learnable projections,
$E_t\!\in\!\mathbb{R}^u$ is a time embedding, and $\phi(\cdot)$ is a
nonlinearity. The alignment loss is defined as
\begin{equation}
\label{eq:lvt}
\mathcal{L}_{\mathrm{FTA}} =
\frac{1}{\mathrm{BDT}}\sum_{b,d,t}
\| \hat{H}^{T}_{b,d,t} - H^{T}_{b,d,t} \|^2 .
\end{equation}

This design bridges the representational gap by factorizing variable identity and temporal modulation. It enables the student to reconstruct point-wise signals from compact variate-wise embeddings, so that temporal knowledge from the teacher can be distilled without inflating the student’s token count.

\subsubsection{Boundary Case Analysis}

\textbf{Proposition.} Factorized Temporal Alignment (FTA) can exactly recover the teacher’s point-wise representations in a boundary case.  

\textbf{Proof.} Let the input sequence of variate $d$ in sample $b$ be $x_{b,d}=[x_{b,d,1},\dots,x_{b,d,T}]$. If the student adopts a \emph{time-separable} representation, its variate-wise embedding is 
\begin{equation}
h^S_{b,d}=[\psi(x_{b,d,1});\dots;\psi(x_{b,d,T})],
\end{equation}
where $\psi$ is a point-wise embedding map without cross-time interaction. FTA generates point-wise embeddings as 
\begin{equation}
z_{b,d,t}=(W_s h^S_{b,d})\odot E_t,\quad \hat H^T_{b,d,t}=W_{\mathrm{out}}\phi(z_{b,d,t}).
\end{equation}
If $E_t$ is one-hot, then $z_{b,d,t}=\psi(x_{b,d,t})$, which exactly selects the $t$-th point embedding. Thus $\hat H^T_{b,d,t}$ aligns with the teacher’s point-wise hidden state. Therefore, the hypothesis space of FTA includes the extreme case of \emph{perfect point-wise alignment}; with learnable $E_t$, the model performs soft time-wise gating, allowing efficient students to approximate point-wise representations without extra tokens.

\subsection{Alternative KD Variants.}
We introduce two simple baselines for comparative experiment.

\textbf{Trend Projection KD (T-KD).} Distills coarse trends via projection:
\begin{equation}
\mathcal{L}_{\text{T-KD}} = \ell(\hat{\mathbf{y}}_s,\mathbf{y}) 
+ \alpha \cdot \ell\!\left(\mathcal{P}(\hat{\mathbf{y}}_s),\mathcal{P}(\hat{\mathbf{y}}_t)\right).
\end{equation}

\textbf{Frequency \& Difference KD (FD-KD).} Aligns spectral and difference features:
\begin{equation}
\mathcal{L}_{\text{FD-KD}} = \ell(\hat{\mathbf{y}}_s,\mathbf{y})
+ \alpha \cdot \big( \beta \cdot \mathcal{L}_{\text{freq}} + \gamma \cdot \mathcal{L}_{\text{diff}} \big).
\end{equation}

Here $\ell$ is MSE.

\section{EXPERIMENT}
\label{sec:exp}

\subsection{Experimental Settings}

\textbf{Datasets.} We conduct extensive experiments on five real-world multivariate time series datasets, including ETTh1, ETTh2, ETTm1, ETTm2~\cite{haoyietal-informer-2021}, and Weather~\cite{liu2023itransformer}, all under the long-term forecasting setting.

\noindent
\textbf{Baseline.} We compare against pre-trained foundation time series models, including TimeMoE (50M, 200M)~\cite{shi2024timemoe}, MOIRAI (small, base, large)~\cite{woo2024moirai}, Chronos (small, base, large)~\cite{ansari2024chronos}, TimesFM~\cite{timesfm}, and Moment~\cite{goswami2024moment}, using the parameter settings as specified in their respective publications.

\noindent
\textbf{Performance metrics.} We evaluate predictive performance using Mean Square Error (MSE) and Mean Absolute Error (MAE). All experiments are repeated five times, and the average results are reported.

\noindent
\textbf{Hyperparameter setting.} The parameter settings of all baselines follow BLAST~\cite{blast} and TimeMoe~\cite{shi2024timemoe}. Since different TSFMs adopt varying lookback windows, the student and teacher models are aligned to the same lookback windows during distillation, to ensure consistent supervision across horizons, since TSFMs are highly sensitive to lookback length. Although our setup follows the widely used BLAST benchmark~\cite{blast}, we exclude horizons of 336 and 720~\cite{bergmeir2024fundamental}, since they extend up to 7.5× longer than the input context. Table ~\ref{tab::long-term} reports the best distilled performance across all teacher TSFMs. For fairness, we omit TimesFM on the Weather dataset that was included in its pretraining corpus.

\noindent
\textbf{Experimental Setup.} All experiments are conducted on a single NVIDIA RTX 3090 GPU with 24GB memory. Models use a hidden dimension of 512 and a feed-forward layer of 2048. For distillation, DLinear students are trained for 10 epochs, while iTransformer students train for 4 epochs, early stopping (patience=3). Inputs are z-score normalized, and MSE is the training objective.

\begin{table*}[t!] 
	\setlength{\tabcolsep}{1.8pt}
	\scriptsize
	\centering
    \caption{Full results of the long-term forecasting task. The prediction lengths are set to {96,192}. The term "\textit{Avg.}" represents the average results across the two prediction lengths. The best and second best outcomes are highlighted in \best{best} and \second{second}, respectively. The notation "$1^{\text{st}}$ \textit{Count}" denotes the frequency of each method achieving the top results.}
	\begin{threeparttable}
		\begin{tabular}{c|c|c c|c c|c c|c c|c c|c c|c c|c c|c c|c c|c c|c c}

			\toprule
			\multicolumn{2}{c}{\multirow{2}{*}{\scalebox{1.1}{Models}}}& \multicolumn{2}{c}{$DistilTS_{L}$} & \multicolumn{2}{c}{$DistilTS_{T}$} & \multicolumn{2}{c}{TimeMoe$_{b}$} & \multicolumn{2}{c}{TimeMoe$_{l}$} & \multicolumn{2}{c}{MOIRAI$_{s}$} & \multicolumn{2}{c}{MOIRAI$_{b}$} & \multicolumn{2}{c}{MOIRAI$_{l}$} & \multicolumn{2}{c}{Chronos$_{b}$} & \multicolumn{2}{c}{Chronos$_{s}$} & \multicolumn{2}{c}{Chronos$_{l}$} & \multicolumn{2}{c}{TimesFM} & \multicolumn{2}{c}{Moment}\\
            
			\multicolumn{2}{c}{} & \multicolumn{4}{c}{\scalebox{0.8}{\textbf{(Ours)}}} & \multicolumn{2}{c}{\scalebox{0.8}{(2025)\cite{shi2024timemoe}}} & \multicolumn{2}{c}{\scalebox{0.8}{(2025)\cite{shi2024timemoe}}} & \multicolumn{2}{c}{\scalebox{0.8}{(2024)\cite{woo2024moirai}}} & \multicolumn{2}{c}{\scalebox{0.8}{(2024)\cite{woo2024moirai}}} & \multicolumn{2}{c}{\scalebox{0.8}{(2024)\cite{woo2024moirai}}} & \multicolumn{2}{c}{\scalebox{0.8}{(2024)\cite{ansari2024chronos}}} & \multicolumn{2}{c}{\scalebox{0.8}{(2024)\cite{ansari2024chronos}}} & \multicolumn{2}{c}{\scalebox{0.8}{(2024)\cite{ansari2024chronos}}} & \multicolumn{2}{c}{\scalebox{0.8}{(2024)\cite{timesfm}}} & \multicolumn{2}{c}{\scalebox{0.8}{(2024)\cite{goswami2024moment}}}\\

			\cmidrule(lr){3-4} \cmidrule(lr){5-6} \cmidrule(lr){7-8} \cmidrule(lr){9-10} \cmidrule(lr){11-12} \cmidrule(lr){13-14} \cmidrule(lr){15-16} \cmidrule(lr){17-18} \cmidrule(lr){19-20} \cmidrule(lr){21-22} \cmidrule(lr){23-24} \cmidrule(lr){25-26} 
			\multicolumn{2}{c}{Metric}& MSE & MAE & MSE & MAE & MSE & MAE & MSE & MAE & MSE & MAE & MSE & MAE & MSE & MAE & MSE & MAE & MSE & MAE & MSE & MAE & MSE & MAE & MSE & MAE \\

			\toprule
			\multirow{3}{*}{\rotatebox[origin=c]{90}{ETTm1}} 
			& 96 & \second{0.304} & 0.352 & \best{0.297} & 0.353 & 0.340 & 0.369 & 0.311 & 0.358 & 0.421 & 0.407 & 0.347 & 0.360 & 0.384 & 0.372 & 0.332 & \second{0.333} & 0.328 & \best{0.332} & 0.457 & 0.403 & 0.361 & 0.370 & 0.654 & 0.527\\
            
			& 192 & \best{0.337} & \best{0.360} & 0.371 & 0.399 & 0.355 & 0.390 & \second{0.350} & 0.383 & 0.431 & 0.421 & 0.378 & 0.381 & 0.410 & 0.391 & 0.384 & \second{0.363} & 0.391 & 0.368 & 0.530 & 0.450 & 0.414 & 0.405 & 0.662 & 0.532 \\
            
			\cmidrule(lr){2-26}
			& \emph{Avg.} & \best{0.320} & 0.356 & 0.334 & 0.376 & 0.347 & 0.379 & \second{0.330} & 0.370 & 0.426 & 0.414 & 0.362 & 0.370 & 0.397 & 0.381 & 0.358 & \best{0.348} & 0.359 & \second{0.350} & 0.493 & 0.426 & 0.387 & 0.387 & 0.658 & 0.529 \\

			\midrule
			\multirow{3}{*}{\rotatebox[origin=c]{90}{ETTm2}} 
			& 96 & \best{0.164} & 0.259 & \second{0.172} & 0.262 & 0.217 & 0.303 & 0.207 & 0.293 & 0.204 & 0.294 & 0.190 & 0.273 & 0.192 & 0.276 & 0.178 & \second{0.245} & 0.174 & \best{0.244} & 0.197 & 0.271 & 0.202 & 0.270 & 0.260 & 0.335 \\
            
			& 192 & \best{0.224} & 0.311 & 0.244 & 0.315 & 0.255 & 0.324 & 0.250 & 0.322 & 0.273 & 0.336 & 0.284 & 0.316 & 0.252 & 0.316 & 0.245 & \best{0.290} & 0.250 & \second{0.296} & \second{0.254} & 0.314 & 0.289 & 0.321 & 0.289 & 0.350 \\
            
			\cmidrule(lr){2-26}
			& \emph{Avg.} &\best{0.194} &0.285 &\second{0.208} &0.288 &0.236 &0.313 &0.228 &0.307 &0.238 &0.315 &0.237 &0.294 &0.222 &0.296 &0.211 &\best{0.267} &0.212 &\second{0.270} & 0.225 &0.292 &0.245 &0.295 &0.274 &0.342\\
            
			\midrule
			\multirow{3}{*}{\rotatebox[origin=c]{90}{ETTh1}} 
			& 96 & 0.362 & 0.390  & {0.374} & 0.403 & \second{0.357} & \second{0.382} & \best{0.350} & \second{0.382}& 0.390 & 0.411 & 0.396 & 0.406 & 0.432 & 0.423 & 0.385 & \best{0.380} & 0.394 & \second{0.382} & 0.441 & 0.390 & 0.414 & 0.404 & 0.688 & 0.557\\
            
			& 192 & 0.398 & \second{0.412} & 0.415 & 0.435 & \best{0.385} & \best{0.405} & \second{0.390} & \second{0.412}  & 0.417 & 0.430 & 0.419 & 0.431 & 0.536 & 0.473 & 0.442 & \second{0.412} & 0.455 & 0.414 & 0.502 & 0.424 & 0.465 & 0.434 & 0.688 & 0.560\\
            
			\cmidrule(lr){2-26} 
			& \emph{Avg.} & 0.380 & 0.401& 0.394 & 0.419 & \second{0.371} & \best{0.394} & \best{0.370} & \second{0.397} &0.403 &0.420 &0.407 &0.418 &0.484 &0.448 &0.413 &0.396 &0.424 &0.398 &0.471 &0.407 &0.439 &0.419 &0.688 &0.558\\
            
			\midrule
			\multirow{3}{*}{\rotatebox[origin=c]{90}{ETTh2}} 
            
			& 96 & \best{0.272} & 0.337  & 0.294 & 0.340 & 0.310 & 0.364 & 0.300 & 0.351 & 0.293 & 0.351 & 0.353 & 0.342 & 0.300 & 0.346 & 0.290 & \second{0.325} & \second{0.283} & \best{0.324} & 0.320 & 0.345 & 0.315 & 0.349 & 0.342 & 0.396\\
            
			& 192 & \best{0.343} & 0.387 & {0.378} & {0.402} & 0.355 & 0.390 & 0.361 & 0.382 & 0.362 & 0.397 & 0.669 & 0.400 & 0.377 & 0.398 & 0.361 & \second{0.368} & 0.371 & 0.377 & 0.406 & 0.399 & \second{0.354} & 0.388 & 0.395 & \best{0.354}\\
            
			\cmidrule(lr){2-26}
			& \emph{Avg.} & \best{0.307} & 0.362 & 0.336 & 0.371 & 0.333 & 0.377 & 0.331 & 0.367 & 0.327 & 0.374 & 0.511 & 0.371 & 0.338 & 0.372 & \second{0.325} & \best{0.346} & 0.327 & \second{0.350} & 0.363 & 0.372 & 0.334 & 0.368 & 0.368 & 0.375\\
            
			\midrule
			\multirow{3}{*}{\rotatebox[origin=c]{90}{Weather}}
            
			& 96  & 0.172 & 0.234 & \best{0.152} & \best{0.205} & \second{0.162} & 0.217 & 0.163 & 0.218 & 0.235 & 0.258 & 0.419 & 0.247 & 0.210 & 0.235 & 0.178 & \second{0.210} & 0.173 & 0.206 & 0.194 & 0.235 & - & - & 0.243 & 0.255 \\
            
			& 192 & 0.211 & 0.281 & \best{0.201} & \second{0.243} & 0.215 & 0.266 & 0.235 & 0.282  & 0.235 & 0.258 & 0.386 & 0.289 & 0.265 & 0.282 & 0.384 & 0.363 & \second{0.212} & \best{0.242} & 0.249 & 0.285 & - & - & 0.278 & 0.329 \\
            
			\cmidrule(lr){2-26}
            
			& \emph{Avg.} & 0.191 & 0.257 & \best{0.176} & \best{0.224} & \second{0.188} & \second{0.241} & 0.199 & 0.250 & 0.235 & 0.258 & 0.402 & 0.268 & 0.237 & 0.258 & 0.281 & 0.286 & 0.192 & \best{0.224} & 0.221 & 0.260 & - & - & 0.260 & 0.292 \\

            
            
            
            \midrule
            \multicolumn{2}{c|}{Total Avg.} & \best{0.278} & 0.332 & 0.289 & 0.335 & 0.295 & 0.341 & \second{0.291} & 0.338 & 0.325 & 0.356 & 0.383 & 0.350 & 0.335 & 0.351 & 0.317 & \second{0.328} & 0.302 & \best{0.318} & 0.354 & 0.351 & 0.351 & 0.367 & 0.449 & 0.419 \\
            
            \midrule
            \multicolumn{2}{c|}{$1^{\text{st}}$ \emph{Count}} & \multicolumn{4}{c|}{\best{13}} & \multicolumn{4}{c|}{5}  & \multicolumn{6}{c|}{0}  & \multicolumn{6}{c|}{5} & \multicolumn{2}{c|}{0} & \multicolumn{2}{c}{0}\\
			\toprule
		\end{tabular}
	\end{threeparttable}
	\label{tab::long-term}
    \vspace{-15pt}
\end{table*}

\begin{figure}[h]
  \centering
  \includegraphics[width=0.95\linewidth]{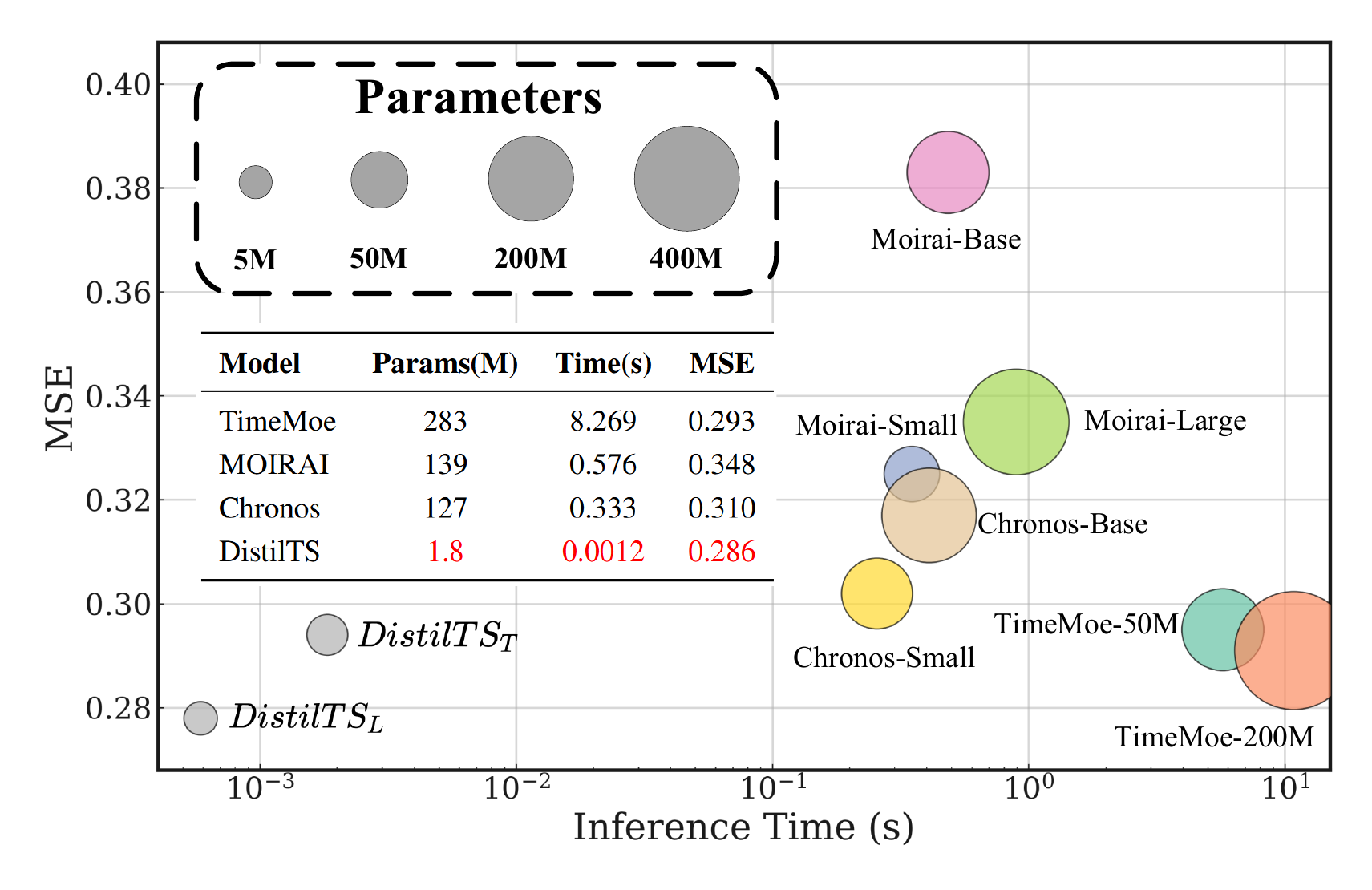}
  \vspace{-12pt}
  \caption{Experiment: Efficiency vs. Performance. The bubble plot presents each model variant separately , while DistilTS is shown as $DistilTS_{L}$ (DLinear~\cite{dlinear} as student) and $DistilTS_{T}$ (iTransformer~\cite{liu2023itransformer} as student). The accompanying table reports the averaged parameter counts and forecasting errors (MSE) within each model family, with DistilTS values averaged over its two variants.}
  \vspace{-13pt}
  \label{fig:eff}
\end{figure}

\subsection{Main Results}

Based on the above setup, Table~\ref{tab::long-term} reports the long-term forecasting results across five benchmarks with prediction lengths of 96 and 192. DistilTS consistently achieves competitive or best performance, ranking first in 13 cases and second in several others. Compared with large TSFMs, our distilled models deliver comparable performance.

Efficiency is a key factor in evaluating forecasting models. As shown in Figure~\ref{fig:eff}, we compare model size (M), inference time (s), and forecasting error across representative TSFMs and DistilTS. While large TSFMs achieve strong performance at the cost of heavy parameters and slow inference, DistilTS reduces the parameter scale by two orders of magnitude and accelerates inference by over three orders of magnitude, while achieving the lowest error. To ensure fairness, parameter counts and inference speed are measured on the ETTh1 dataset, to decouple efficiency and forecasting robustness for a fairer evaluation, whereas forecasting performance is reported as the average MSE across all five benchmark datasets.



\begin{figure}[h]
  \centering
  \includegraphics[width=0.95\linewidth]{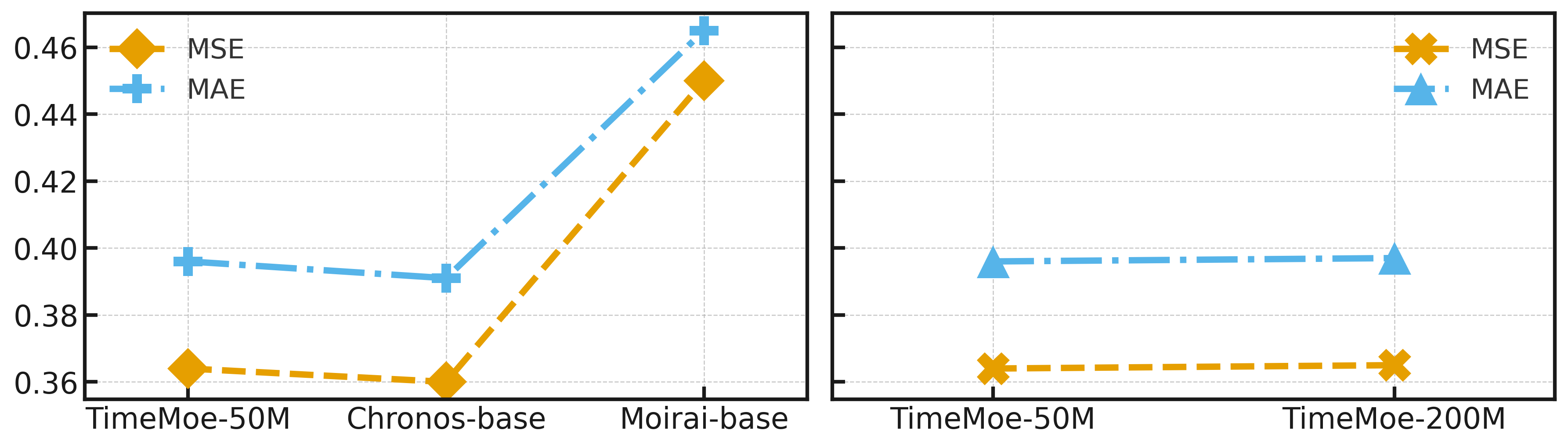}
  \caption{Ablation studies of DistilTS under different teacher settings. Left: Comparison across different TSFMs as teachers, showing that the choice of teacher architecture significantly impacts distillation performance. Right: Comparison within the same architecture (TimeMoe) at different scales (50M vs. 200M), showing that increasing teacher size provides only limited additional benefits.}
  \vspace{-10pt}
  \label{fig:etth1_dual}
\end{figure}

\subsection{Ablation Experiments}

To further analyze the DistilTS, we conduct ablation studies. Figure~\ref{fig:etth1_dual} reports the distillation performance of DLinear on ETTh1. Across different TSFMs, DistilTS achieves competitive errors, with TimeMoe and Chronos outperforming Moirai. When comparing TimeMoe-50M and TimeMoe-200M, suggesting that beyond a certain scale, teacher size contributes marginal gains to distillation. The parameter scale of teacher models has limited impact on student performance, likely because TSFMs do not strictly follow scaling laws; in fact, larger models may even degrade forecasting quality, as evidenced in Table 5 of BLAST~\cite{blast}. $DistilTS_{L}$ denotes our method distilled with DLinear~\cite{dlinear} as the student baseline, and $DistilTS_{T}$ denotes our method distilled with iTransformer~\cite{liu2023itransformer} as the student baseline. Both variants are distilled from the same teacher. The results in Table ~\ref{tab:abla} clearly demonstrate that DistilTS consistently improves upon its baselines. Table~\ref{tab:etth2} illustrates that DistilTS consistently outperforms both FD-KD and KD on ETTh2, achieving lower errors in terms of both MSE and MAE.

\begin{table}[h]
\footnotesize
\centering
\vspace{-10pt}
\caption{Ablation results (MSE) with TimeMoe-50M as teacher.}
\vspace{3pt}
\setlength{\tabcolsep}{2pt}

\renewcommand{\arraystretch}{1.15}
\begin{tabular}{c|c|cccc}
\toprule
\textbf{Dataset} & \textbf{Pred} 
& \textbf{DistilTS} & \textbf{Only HW} & \textbf{Only FTA} & \textbf{Baseline} \\
\midrule
\multirow{3}{*}{\rotatebox[origin=c]{90}{ETTh1}}
& 96  & \textbf{0.374} & 0.382 & 0.378 & 0.386 \\
& 192 & \textbf{0.415} & 0.421 & 0.419 & 0.441 \\
& Avg & \textbf{0.395} & 0.402 & 0.399 & 0.414 \\
\midrule
\multirow{3}{*}{\rotatebox[origin=c]{90}{ETTm2}}
& 96   & \textbf{0.172} & 0.176 & 0.174 & 0.180\\
& 192  & \textbf{0.244} & 0.246 & 0.246 & 0.250\\
& Avg  & \textbf{0.208} & 0.211 & 0.210 & 0.215 \\
\bottomrule
\end{tabular}
\vspace{-10pt}
\label{tab:abla}
\end{table}

\begin{table}[h]
\small
\centering
\caption{Comparison on different distillation object.}
\setlength{\tabcolsep}{5pt}
\begin{tabular}{c|ccc}
\toprule
Method & $DistilTS_{L}$ & FD-KD & T-KD \\
\midrule
MSE/MAE & \textbf{0.287}/\textbf{0.355} & 0.290/0.357& 0.291/0.358 \\
\bottomrule
\end{tabular}
\label{tab:etth2}
\vspace{-15pt}
\end{table}

\section{Conclusion}
\label{sec:conclusion}

We introduced DistilTS, the first distillation framework specifically designed for time-series foundation models (TSFM). By addressing key challenges of horizon imbalance and variate-temporal mismatch, DistilTS enables lightweight student models that retain the forecasting strength of large teacher models. Experimental results demonstrate that DistilTS achieves performance comparable to full-sized TSFMs while delivering orders-of-magnitude improvements
in efficiency.

\clearpage


\bibliographystyle{IEEEbib}
\bibliography{refs}

\end{document}